\def\year{2018}\relax
\documentclass[letterpaper]{article} 
\usepackage{aaai18}  
\usepackage{times}  
\usepackage{helvet}  
\usepackage{courier}  
\usepackage{url}  
\usepackage{graphicx}  
\usepackage{amsmath}
\usepackage{amssymb}
\usepackage{enumitem}
\usepackage{multirow}
\begin{document}
%
\title{An Unsupervised Model with Attention Autoencoders for Question Retrieval}
\author{Minghua Zhang, Yunfang Wu\thanks{Corresponding author} \\
Key Laboratory of Computational Linguistics, Peking University, MOE, China\\
{\tt \{zhangmh, wuyf\}@pku.edu.cn}
}
\maketitle
\begin{abstract}
Question retrieval is a crucial subtask for community question answering. Previous research focus on supervised models which depend heavily on training data and manual feature engineering. In this paper, we propose a novel unsupervised framework, namely reduced attentive matching network (RAMN), to compute semantic matching between two questions. Our RAMN integrates together the deep semantic representations, the shallow lexical mismatching information and the initial rank produced by an external search engine. For the first time, we propose attention autoencoders to generate semantic representations of questions. In addition, we employ lexical mismatching to capture surface matching between two questions, which is derived from the importance of each word in a question. We conduct experiments on the open CQA datasets of SemEval-2016 and SemEval-2017. The experimental results show that our unsupervised model obtains comparable performance with the state-of-the-art supervised methods in SemEval-2016 Task 3, and outperforms the best system in SemEval-2017 Task 3 by a wide margin.
\end{abstract}

\section{Introduction}
\label{sect:introduction}

Community question answering (CQA) portals, like Yahoo! Answers and Baidu Knows, are popular forums where users ask and answer questions on diverse topics. As a consequence, the CQA has accumulated a large quantity of questions and answers, which have made CQA portals valuable resources. In these CQA forums, questions may be repeated or closely related to previously asked questions, and there exists a large amount of answers with respect to a given question. As a result, it will be difficult and time-consuming for users to search and distinguish the good answers. The CQA system could help to automate the process of finding good answers to new questions in a community-created discussion forum. The system first retrieves similar questions (i.e. question retrieval) in the forum and then identifies the posts (i.e. answer selection) in the answer threads of those similar questions. In this paper, we focus on question retrieval for CQA, with the purpose of developing semantic textual similarity measures for such noisy texts.

The challenge of question retrieval is that two natural language sentences often express similar meanings with different but semantically related words, which results in semantic gaps between them. In previous work, various approaches have been proposed to bridge the semantic gaps between two objects, most of which are supervised. Some researches leverage word-to-word (or phrase-to-phrase) translation probabilities to capture sematic matching terms \cite{xue2008retrieval,zhou2011phrase}. In order to train the translation model, one needs to collect a large amount of monolingual parallel strings of similar questions pairs, which are generally not available in practice. Alternatively, question-answer pairs are also treated as parallel strings, but the assumption that questions and answers are semantically equivalent is hardly true in reality. Furthermore, many research \cite{romeo2016neural,barron2016selecting,franco2016uh,charlet2017simbow} apply a learning to rank architecture to handle question retrieval. These methods are supervised fusion of different features, including both supervised and unsupervised similarity features. Among the features, many were based on overlap count between components which include but are not limited to, n-grams of words or characters, named entities, frame representations and knowledge graphs \cite{franco2016uh}. Recently, much attention is also paid for the use of neural matching features \cite{goyal2017learningtoquestion}, which can extract high level matching signals from distributed representations of the sentences and capture their similarity beyond lexicons. Overall, the limitation of previous work is that, building a large amount of training data with similar questions is expensive and careful feature engineering is time-consuming.

In this paper, we propose an unsupervised framework to compute question-question similarity, relying only on the large collection of unannotated data in CQA forums. We focus on core textual semantic similarity, avoiding using any metadata analysis (such as user profile and question categories). We propose a reduced attentive matching network (RAMN), which integrates three aspects of information in a robust and simple unsupervised framework, including the question representations generated by the deep network, the lexical mismatching information based on the surface matching and the initial rank produced by an external search engine. We conduct experiments on the benchmark CQA datasets of Semeval-2016 \cite{nakov2016semeval} and Semeval-2017 \cite{nakov2017semeval}. Evaluation results show that our unsupervised model outperforms the winner system at the campaign, and obtains comparable results with state-of-the-art methods which are all supervised. Our model is unsupervised and domain independent, and so can be easily generalized to other text-matching tasks, like answer selection and paraphrase detection.

First, we propose attention autoencoders to embed questions into semantic representations, which is pre-trained using a large scale unannotated data. In the recurrent neural network (RNN) architecture, those representations near the end of a sentence are likely to contain more information, which may result in biased representations towards the end of a sentence. Our attention autoencoders is inspired by the work of \citeauthor{Vaswani2017Attention} \shortcite{Vaswani2017Attention}, with the goal of generating the input sequence itself. The representation from attention autoencoders contains context information with a strong focus on the current word of the input sequence, which is more suitable for semantic matching. What's more, the attention autoencoders allow for more parallelization and achieve significant improvements in computational efficiency.

Recently, researchers have proposed various neural network models to deal with question answering. The strong generalization power enables these methods to find texts with similar latent representations, but they may miss or obscure the detailed matching information. In this paper, we employ mismatching to find exact matching in question retrieval. In practice, if there is a key term in the new question not appearing in the candidate question, the similarity between them should be reduced. In contrast, the similarity should be less vulnerable if a background word of the new question does not appear in the candidate question. Following the observation, we propose a simple but effective method to capture the lexical mismatching information. We automatically calculate the importance of each word in the new question, and then acquire a reduced vector according to lexical mismatching relation between the new question and candidate question.

We summarize our contributions in this paper as follows:
\begin{itemize}
\item We propose a new unsupervised architecture RAMN for question retrieval, in which the deep question similarity, the lexical mismatching score and the external searching rank are seamlessly integrated.
\item For the first time, we propose attention autoencoders to generate sentence representations in an unsupervised manner.
\item We propose to model surface matching by computing lexical mismatching information in an unsupervised way, which obviously improves the performance.
\end{itemize}

\section{Problem Formalization}
\label{sect:formalization}

Suppose that we have a new question $q$ and a candidate question $Q$. Formally, we denote the embedding sequence of $q$ as $\{ x_{qi} | 0<i<=n, i \in N_+\}$, the embedding sequence of $Q$ as $\{ x_{Qj} | 0<j<=m, j \in N_+\}$, where $n$ and $m$ are the number of words in $q$ and $Q$ respectively. $h_{qi}$ and $h_{Qj}$ are hidden representations of $q$ at step $i$ and $Q$ at step $j$ respectively. Based on the lexical mismatching relation between $q$ and $Q$, we can compute a reduced vector which we represent as $d_q$. Our goal is to design a matching model $g(\cdot, \cdot)$ with hidden representations and the reduced vector. For each question pair $(q, Q)$, $g(q, Q)$ returns a matching score which can be used to rank candidate questions for the new question.

To obtain $g(q, Q)$, we need to answer two questions: 1) how to represent questions in the latent space using unsupervised methods, and 2) how to capture surface matching information and incorporate it into the matching model. In the following sections, we first present our method on question representation, and then elaborate on our matching model.

\section{Attention Autoencoders for Question Representation}
\label{sect:autoencoder}

The sequence autoencoder \cite{dai2015semi} is similar to sequence to sequence learning (also known as seq2seq) \cite{sutskever2014sequence}. It employs a recurrent network as an encoder to read in an input sequence into a hidden representation. Then, the representation is fed to a decoder recurrent network to reconstruct the input sequence itself. The sequence autoencoder is an unsupervised learning model which is a powerful tool for modeling sentence representations with large scale unannotated data.

Different from the traditional RNN autoencoders, we propose attention autoencoders to model sentence representations. Our approach is inspired by the work of neural sequence transduction models \cite{Vaswani2017Attention}, which has been successfully applied for machine translation. It is the first transduction model relying entirely on self-attention to compute representations of the input and output sequences without using RNNs or convolution. Our autoencoder is similar to the concept, except that it is an unsupervised learning model and the objective is to reconstruct the input sequence itself. The experimental results show that our attention autoencoders obtain better performance over RNN autoencoders in reconstructing the input sequence.

\begin{figure}
\centering
\includegraphics[scale=1.3]{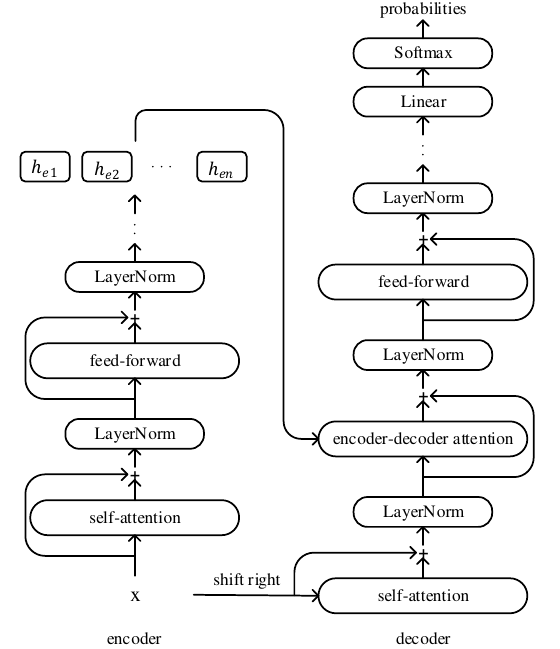}
\caption{\label{autoencoder} The attention autoencoders applied in our model.}
\end{figure}

Our attention autoencoders follow the encoder-decoder architecture, as shown in Figure \ref{autoencoder}. Each layer of encoder has two sub-layers: the first layer is a self-attention mechanism, and the second is a position-wise feed-forward network. Besides the two sub-layers, the decoder inserts a third sub-layer, which performs attention over the output of the encoder. In the self-attention sub-layer of the decoder, we apply mask to ensure that the predictions for position $i$ can depend only on the known outputs at positions less than $i$. We also employ residual connection and layer normalization around each of the sub-layers.

The attention mechanism is to map a query and a set of key-value pairs to an output. The output is computed as a weighted sum of the values, where the weight assigned to each value is computed based on the query and the corresponding key. In attention autoencoders, there are three types of attention: the encoder self-attention, the encoder-decoder attention and the decoder self-attention. We compute the attention on a set of queries simultaneously:
\begin{align}
\label{equ att}
& Attention(Q, K, V) = softmax( \frac{Q K^T}{ \sqrt{d_k} } ) V
\end{align}
where $Q \in \mathbb{R}^{n_q \times d_k}$, $K \in \mathbb{R}^{n_k \times d_k}$ and $V \in \mathbb{R}^{n_k \times d_v}$ are queries, keys and the corresponding values respectively; $n_q$ is the number of queries and $n_k$ is the number of key-value pairs.

The encoder transforms a sentence into a list of vectors, one vector per input symbol. Given the input embedding sequence $\mathbf{x} = (x_1, \dots, x_n)$, we produce hidden representations $\mathbf{h_e} = (h_{e1}, \dots, h_{en})$ with the following equations:
\begin{align}
\label{equ eatt}
& \mathbf{a^\prime_e} = Attention(\mathbf{x} W^q_e, \mathbf{x} W^k_e, \mathbf{x} W^v_e) \\
\label{equ eattln}
& \mathbf{a_e} = LayerNorm(\mathbf{a^\prime_e} + \mathbf{x}) \\
\label{equ eff}
& \mathbf{h^\prime_e} = ReLU(\mathbf{a_e} W_{e1} + b_{e1}) W_{e2} + b_{e2} \\
\label{equ effln}
& \mathbf{h_e} = LayerNorm(\mathbf{h^\prime_e} + \mathbf{a_e})
\end{align}
where $W^q_e \in \mathbb{R}^{d_m \times d_k}$, $W^k_e \in \mathbb{R}^{d_m \times d_k}$, $W^v_e \in \mathbb{R}^{d_m \times d_e}$, $W_{e1} \in \mathbb{R}^{d_m \times d_f}$ and $W_{e2} \in \mathbb{R}^{d_f \times d_m}$ are parameter matrices; $b_{e1} \in \mathbb{R}^{d_f}$ and $b_{e2} \in \mathbb{R}^{d_m}$ are bias vectors; $LayerNorm$ denotes layer normalization and $ReLU$ is activation function.

Given the encoder representations $\mathbf{h_e}$, the decoder is responsible for generating the input sequence. The encoder and decoder are connected through an attention module, which allows the decoder to focus on different parts of the input sequence during the course of decoding.

We first shift the input embedding $\mathbf{x}$ right and get $\mathbf{x^\prime} = (0, x_1, \dots, x_{n-1})$ as the decoder input. Through employing equation \eqref{equ eatt} and \eqref{equ eattln} in the decoder self-attention layer, we get $\mathbf{a_d} = (a_{d1}, \dots, a_{dn})$. The encoder-decoder attention is applied following the self-attention layer, which is computed as:
\begin{align}
\label{equ deatt}
& \mathbf{a^\prime_{ed}} = Attention(\mathbf{a_d} W^q_a, \mathbf{h_e} W^k_a, \mathbf{h_e} W^v_a) \\
& \mathbf{a_{ed}} = LayerNorm(\mathbf{a^\prime_{ed}} + \mathbf{a_d})
\end{align}
where $W^q_a \in \mathbb{R}^{d_m \times d_k}$, $W^k_a \in \mathbb{R}^{d_m \times d_k}$ and $W^v_a \in \mathbb{R}^{d_m \times d_m}$ are parameter matrices in encoder-decoder attention layer.

Then, $\mathbf{a_{ed}}$ is fed to the position-wise feed-forward layer to produce the hidden representations $\mathbf{h_d} = (h_{d1}, \dots, h_{dn})$. Given $\mathbf{h_d}$ and the previous $(i-1)$ words, the probability of generating word $w_i$ is:
\begin{equation}
\label{equ pw}
P(w_i|w_1,\dots,w_{i-1},h_{di}) \propto exp(W^p h_{di} + b^p)
\end{equation}

The objective is the sum of the log-probabilities for the input sequence itself:
\begin{equation}
\label{equ loss}
J(\theta) = {\sum_{i}{log P(w_i|w_1,\dots,w_{i-1},h_{di})}}
\end{equation}

The attention autoencoders learns to reconstruct the input sequence by optimizing the objective in equation \eqref{equ loss}. In CQA forums, most of questions are unlabeled, and only a small fraction of questions are labeled manually for research usage. The attention autoencoders are very suitable for us to make better use of the unlabeled data in CQA archives.

\section{Reduced Attentive Matching Network}
\label{sect:network}

\begin{figure}
\centering
\includegraphics[scale=1.3]{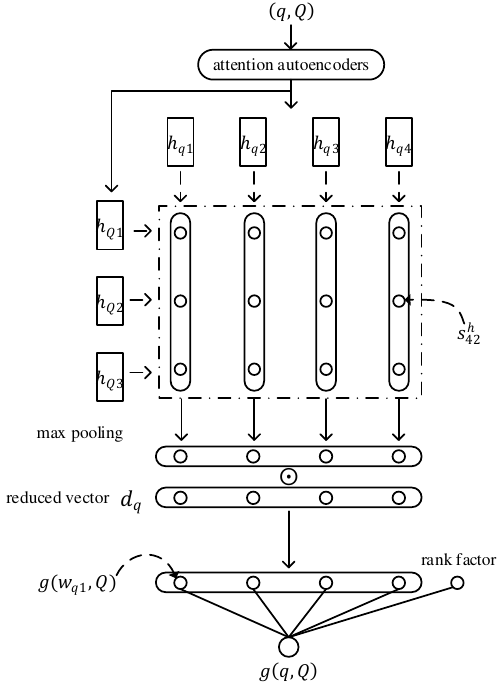}
\caption{\label{framework} The overall structure of our model. A circle denotes a real number, $\odot$ is element-wise product.}
\end{figure}

Based on the question representations generated by attention autoencoders, we further propose a reduced attentive matching network (RAMN) to handle question retrieval. Figure \ref{framework} gives the architecture of our model. Given a new question $q$ and a candidate question $Q$, our model first embeds them into sequences of hidden representations by our attention autoencoders, and computes the deep semantic matching vector. Further, we compute lexical mismatching representation to capture the surface matching information. Finally, we incorporate the initial rank produced by a search engine into our model.

\subsection{Similarity based on Hidden Representations}
\label{ssec:hidden}

After obtaining the hidden representation $H^q = \{h_{qi} | 0<i<=n\}$ of new question $q$ and the hidden representation $H^Q = \{h_{Qj} | 0<j<=m\}$ of candidate question $Q$, we compute the interactions of each paired segments in $q$ and $Q$. Specifically, for the $i$-th hidden state $h_{qi}$ in $q$ and the $j$-th hidden state $h_{Qj}$ in $Q$, their interaction $s_{ij}^h$ is calculated by:
\begin{equation}
\label{equ hdot}
s_{ij}^h = h_{qi} \cdot h_{Qj}
\end{equation}

In order to compute the matching score between $q$ and $Q$, we first use row-wise pooling to obtain a vector that summarizes the interaction of each segment in query with all segments in candidate question. Then, we decomposes the query-question similarity $g(q, Q)$ to a product of word-question similarities $g(w_{qi}, Q)$. Thus, our base model can be formulated as:
\begin{align}
\label{equ gwQ}
& g(w_{qi}, Q) = \max_{j} s_{ij}^h \\
\label{equ gqQ}
& g(q, Q) = \prod_{i} g(w_{qi}, Q)
\end{align}

In our model, in contrast to the one-hot representation, the hidden representations are able to capture semantic relations of different words effectively, and can partially overcome the lexical gaps. Moreover, we can make use of the large scale unannotated data in CQA to learn effective hidden representations of questions, and we needn't worry about how to collect lots of annotated data or monolingual parallel strings.

\subsection{Lexical Mismatch}
\label{ssec:mismatch}

The match between hidden representations can capture semantic relations between two questions. However, it ignores lexical matching which is considered as one of the important features for question retrieval. Therefore, we propose to model lexical matching through lexical mismatching.

We first calculate the importance of each word in a question through a simple and effective approach. In information retrieval, the TF-IDF is a classical algorithm to compute term weighting, which reflects how important a word $w$ is to a document $D$ in a large corpus $C$:
\begin{equation}
\label{equ tfidf}
TF-IDF(w|D,C) = tf(w|D) idf(w|C)
\end{equation}

Inspired by the TF-IDF, we calculate the frequency of a word in corpus $C$, and the importance of a word $w_{qi}$ in the new question $q$ is given by:
\begin{equation}
\label{equ qf}
f_{qi} = \frac {tf(w_{qi}|C)}{\sum_{ w_{qi^\prime} \in q} tf(w_{qi^\prime}|C)}
\end{equation}
The smaller $f_{qi}$ is, the more important $w_{qi}$ is in $q$. Such a definition allows us to model lexical matching in a more preferable way.

Based on the importance of each word, we can obtain the reduced vector of $q$ which we represent as $d_q$. If $w_{qi}$ exists in the candidate question, we set $d_{qi}$ to one. Otherwise, $d_{qi}$ will be equal to $f_{qi}$, which implies reduced match.

For example, given a $q$ and a $Q$:
\begin{itemize}[leftmargin=2.6em]
\item[\textbf{q:}] \textit{We propose an unsupervised model}
\item[\textbf{Q:}] \textit{We propose a supervised model}
\end{itemize}

The importance of each word is given in \eqref{equ qfv}, and then $d_q$ can be obtained as in \eqref{equ qD}:
\begin{align}
\label{equ qfv}
& (0.5401,0.0075,0.4221,0.0008,0.0295) \\
\label{equ qD}
& d_q = (1,1,0.4221,0.0008,1)
\end{align}
where the query words ``\textit{We}'', ``\textit{propose}'' and ``\textit{model}'' exist in $Q$, and we set the corresponding elements in $d_q$ to one. ``\textit{an}'' and ``\textit{unsupervised}'' don't appear in $Q$, and $d_{q3}$ and $d_{q4}$ are equal to $f_{q3}$ and $f_{q4}$ respectively. It demonstrates that, ``\textit{unsupervised}'' is the most important word in $q$, the mismatch of which will result in a serious penalization in similarity computation.

Equipped with the reduced vector, we can overwrite the similarity computation in equation \eqref{equ gwQ} by:
\begin{equation}
\label{equ gqQ D}
g(w_{qi}, Q) = d_{qi} \max_{j} s_{ij}^h
\end{equation}
So, the more important $w_{qi}$ is in $q$, the more serious reduce $g(w_{qi}, Q)$ suffer when $w_{qi}$ mismatches in $Q$.

\subsection{Rank Factor}
\label{ssec:rank}

In most cases, the research goal of question retrieval is to re-rank candidate questions initially ranked by a search engine. The initial rank is often computed using powerful algorithms, so it is essential to incorporate it into the matching model. For each candidate question $Q$, we compute a rank factor $R$ using the equation \eqref{equ rank} where $\alpha$ is a parameter which is tuned in the development set. We apply $R$ to update equation \eqref{equ gqQ} as in \eqref{equ gqQ R}:
\begin{align}
\label{equ rank}
& R = 1 - \alpha * rank \\
\label{equ gqQ R}
& g(q, Q) = R \prod_{i} g(w_{qi}, Q)
\end{align}

\section{Experiments}
\label{sect:experiment}

\subsection{Dataset}
\label{ssec:data}

We conduct experiments on the CQA datasets of SemEval-2016 Task 3 and SemEval-2017 Task 3. These datasets contain real data from the community-created Qatar Living Forums. There are three English subtasks and we focus on Subtask B: Question-Question Similarity. The SemEval-2017 Task 3 is an extended edition of SemEval-2016 Task 3, where the organizers reuse the same training dataset from SemEval-2016 but annotate fresh test sets for all subtasks. For a new question, there are 10 candidate questions retrieved by the Google search engine. The research goal is to re-rank the candidate questions according to their similarity with respect to the new question.

The task is in a supervised setting. Human annotators were asked to assign one of the three labels to each candidate question: perfectMatch, relevant and irrelevant. PerfectMatch and relevant questions are regarded as positive instances, and irrelevant questions as negative instances. The labeled dataset is divided into three folders: training, development and test. Table \ref{table dataset} gives the statistics distribution of the dataset.

Our model is unsupervised, and so we only use the development data to tune parameters, and use the same test data to compare our methods with previous research.  In our model, any external resources of unlabeled data can be utilized to train attention autoencoders. In our experiment, we utilize the large amount of unlabeled data released by the organizers, which consists of 189,941 questions and 1,894,456 comments.

In this task, the evaluation metrics are MAP and MRR, which are widely used for question retrieval. We utilize the official evaluation script published by the organizer.

\begin{table}
\centering
\begin{tabular}{lcccc}
\hline
Class & Train & Dev & 2016-Test & 2017-Test \\
\hline
\hline
Original & 267 & 50 & 70 & 88 \\
\hline
Candidates & 2669 & 500 & 700 & 880 \\
PerfectMatch & 235 & 59 & 81 & 24 \\
Relevant & 648 & 155 & 152 & 139 \\
Irrelevant & 1586 & 286 & 467 & 717 \\
\hline
\end{tabular}
\caption{\label{table dataset} Statistics distribution in the training, development and test partitions on the Subtask B.}
\end{table}

\subsection{Experimental Setup}
\label{ssec:setup}

We concatenated the subject and main body of a question to be a unique question. All texts were tokenized, lowercased and stemmed to reduce the dimensionality of the dictionary. For computational reasons, we opted to limit the size of the input texts at 128 words, and we did not observe any degradation in performance.

For the attention autoencoders, the dimensionality of word embeddings was set to 200. Word embeddings were initialized by the result of word2vec \cite{mikolov2013distributed} trained on unannotated Qatar data and updated in training. Tokens that did not appear in the pre-trained word embeddings were replaced with UNK symbol, of which the embeddings were initialized randomly.

In training the autoencoders, we initialized all attention parameters with orthogonal initialization. The parameters of position-wise feed-forward layers were initialized by a uniform distribution in $[-0.1,0.1]$. Both the encoder and decoder consist of a stack of 2 identical layers. The dimension of hidden representation was set to 200, which was equal to the dimensionality of word embeddings. We applied the Adam algorithm \cite{kingma2014adam} to optimize the attention autoencoders, using shuffled mini-batches of size 48. The initial learning rate is 0.0004. The autoencoders learn until the performance in the development data stops improving, with patience $p = 3$, i.e., the number of epochs to wait before early stopping.

The unlabeled data used in attention autoencoders also served as the corpus $C$ in equation \eqref{equ qf}. We set $\alpha$ in equation \eqref{equ rank} to 0.035, which was tuned via grid search over the following range of values $\{0.01,0.015,0.02,\dots,0.1\}$.

\subsection{Baselines}
\label{ssec:baselines}

We compare our model with the following state-of-the-art baselines, which are all supervised methods.

\begin{itemize}
\item \noindent{\bf SemEval-2016 Best} \cite{franco2016uh}: This method employed Support Vector Machines (SVMrank) to rank candidate questions, by using a variety of lexical and semantic features.
\item \noindent{\bf Tree Pruning} \cite{romeo2016neural}: It employed LSTM with an attention mechanism to select the important sentences as well as the important chunks/words from syntactic parsing trees, and then exploited implicit features of the subtrees to re-rank questions.
\item \noindent{\bf Tree kernel classifier} \cite{barron2016selecting}: It also addressed the problem of selecting the most relevant text chunks in the questions. It employed supervised and unsupervised models that operated both at sentence and chunk level (using constituency trees). A tree-kernel-based classifier was utilized.
\item \noindent{\bf SemEval-2017 Best} \cite{charlet2017simbow}: The approach was a supervised combination of different unsupervised textual similarities, by introducing a soft-cosine that takes into account relations between words. The features were fused using logistic regression.
\item \noindent{\bf SemEval-2017 Second} \cite{goyal2017learningtoquestion}: It employed pairwise learning to rank methods, and used various semantic features to achieve promising results on this subtask.
\item \noindent{\bf SemEval-2017 Third} \cite{filice2017kelp}: It modeled the task as a binary classification problem. All classifiers and kernels were implemented in the Kernel-based Learning Platform (KeLP). It is the only system that appears in the top-3 ranking for all the three English subtasks.
\end{itemize}

\subsection{Evaluation Results}
\label{ssec:eval}

The experimental results are shown in Table \ref{table best} for SemEval-2016 and SemEval-2017.

\begin{table}
\centering
\begin{tabular}{clcc}
\hline
 & Model & MAP & MRR \\
\hline
\hline
\multirow{5}{*}{2016} & Baseline (IR) & 74.57 & 83.79 \\
 & SemEval Best & 76.70 & 83.02 \\
 & Tree Pruning & 77.82 & 84.64 \\
 & Tree Kernel Classifier & 78.56 & 85.12 \\ \cline{2-4}
 & Our model & 77.79 & 85.76 \\
\hline
\hline
\multirow{5}{*}{2017} & Baseline (IR) & 41.85 & 46.42 \\
 & SemEval Best & 47.22 & 50.07 \\
 & SemEval Second & 46.93 & 53.01 \\
 & SemEval Third & 46.66 & 50.85 \\ \cline{2-4}
 & Our model & 48.53 & 52.75 \\
\hline
\end{tabular}
\caption{\label{table best} Experimental results of our model comparing with the state-of-the-art methods.}
\end{table}

At SemEval-2016, our model performs best over all state-of-the-art methods in term of MRR. And our model also obtains a promising MAP score of 77.79, which outperforms the strong baseline by 3.22 and beats the best system by 1.09. Our result is comparable with the Tree Pruning method, and is a little lower (less than 1 MAP point) than the Tree Kernel Classifier. At SemEval-2017, our model outperforms the best system with 1.31 MAP points, and beats the baseline by a wide margin (6.68 MAP points).

It is worth noticing that the MAP score adopted in this campaign is sensitive to the amount of original questions which don't have any relevant questions in the gold labels. In fact, these questions always account for a precision of 0 in the MAP scoring. Comparing the two subtasks of SemEval-2016 and SemEval-2017, the SemEval-2017 Task 3 is more challenging. The upper bound of MAP performance is 88.57 for SemEval-2016, but only 67.05 for SemEval-2017; the baseline is 74.57 for SemEval-2016, but only 41.85 for SemEval-2017. Our model obtains much gain at SemEval-2017, demonstrating that it is more robust to the hard task.

It should be noted again that our model is unsupervised, while the above state-of-the-art methods are all supervised and need training data to train the model. As everyone know, to construct a training data is time consuming, laborious and expensive. Our model needs only a large amount of unlabeled data to train autoencoders, which can be easily downloaded from CQA forums. What’s more, it is a non-trivial task for supervised methods to do domain adaptation.

Our method is in an end-to-end manner, while the above state-of-the-art methods need rich features and external resources. They did careful feature engineering work, and generated features from external knowledge bases (like WordNet and FrameNet). In addition, they needed a sophisticated parser to effectively parse the full text. \citeauthor{romeo2016neural} \shortcite{romeo2016neural} conducted an end-to-end LSTM network to treat question ranking as a classification task, but the experimental result was poor with a MAP of 67.96. As they reported, the small dataset (only 2,669 training examples) is not qualified with complicated neural network methods.

\subsection{Ablation Study for Model Components}
\label{ssec:analysis}

To further analyze our model especially the effectiveness of lexical mismatch, we make a detail analysis on each step of the model, by removing one component at a time. Table \ref{table step} report the evaluation results at SemEval-2016 and SemEval-2017, where the base model refers to the semantic matching based only on the question representations generated by attention autoencoders.

\begin{table}
\centering
\begin{tabular}{clcc}
\hline
 & Models & MAP & MRR \\
\hline
\hline
\multirow{4}{*}{2016} & SemEval Best & 76.70 & 83.02 \\ \cline{2-4}
 & Our base model & 73.05 & 81.51 \\
 & + lexical mismatch & 75.92 & 84.29 \\
 & + initial rank & 77.79 & 85.76 \\
\hline
\hline
\multirow{4}{*}{2017} & SemEval Best & 47.22 & 50.07 \\ \cline{2-4}
 & Our base model & 45.36 & 51.27 \\
 & + lexical mismatch & 47.40 & 51.61 \\
 & + initial rank & 48.53 & 52.75 \\
\hline
\end{tabular}
\caption{\label{table step} Experimental results of our model with respect to each step.}
\end{table}

In our model, the performance is consistently increased step by step at both datasets. The performance is obviously improved when introducing the lexical mismatch into the model, where the improvement over the base model is 2.87 MAP points at SemEval-2016 and 2.04 points at SemEval-2017. In this step, comparing with the best system at the campaign, our model obtains comparable results at SemEval-2016 and better results at SemEval-2017. The rank factor further improves 1.87 and 1.13 respectively in two datasets.

It shows that the proposed unsupervised matching model, which fuses the distributed representations, the lexical mismatching information and the initial rank, is extremely robust and effective in tackling learning tasks defined on sentence pairs.

\subsection{Analysis on Attention Autoencoders}
\label{ssec:compare}

\begin{table}
\centering
\begin{tabular}{cccc}
\hline
 & Autoencoders & Base model & Full model \\
\hline
\hline
\multirow{2}{*}{2016} & RNN & 71.76/81.14  & 77.06/84.16 \\
 & Attention & 73.05/81.51 & 77.79/85.76 \\
\hline
\hline
\multirow{2}{*}{2017} & RNN & 43.59/48.80 & 46.38/52.07 \\
 & Attention & 45.36/51.27 & 48.53/52.75 \\
\hline
\end{tabular}
\caption{\label{table rnn}  Experimental results (MAP/MRR) of attention autoencoders comparing with RNN autoencoders.}
\end{table}

\begin{figure}
\centering
\includegraphics[scale=0.052]{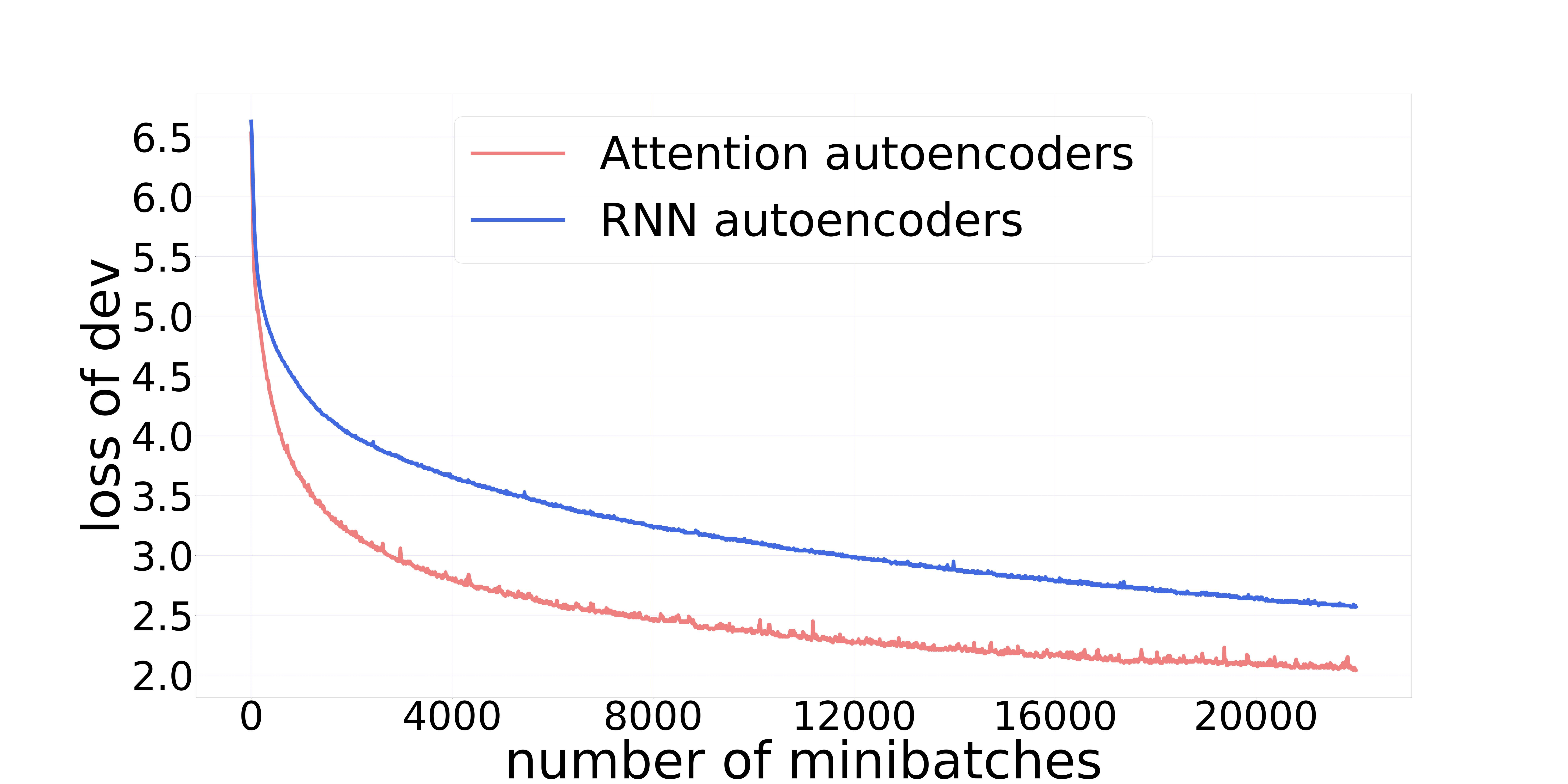}
\caption{\label{loss} The training loss curve of RNN autoencoders and attention autoencoders.}
\end{figure}

In this paper, we propose attention autoencoders and apply them to question retrieval. Here, we conduct experiments to compare our attention autoencoders with the traditional RNN autoencoders. We create a RNN autoencoder which consists of a two-layer encoder with GRU activation and a two-layer decoder with conditional GRU. In the encoder, we produce sentence representation by average pooling over all hidden representations. Then, the sentence representation is fed to the decoder to reconstruct the input sequence itself. The parameter configurations are consistent with our attention autoencoder.

Table \ref{table rnn} report the evaluation results at SemEval-2016 and SemEval-2017. The base model refers to the semantic matching based only on the question representations generated by RNN/attention autoencoders, and further integrating the lexical mismatch information and rank factor forms the full model.

At both SemEval-2016 and SemEval-2017, the semantic matching with attention autoencoders obtains better performance than the RNN autoencoders. It shows more clearly in the base model, where the attention autoencoders outperform the RNN autoencoders by 1.29 MAP points at SemEval-2016 and 1.77 MAP points at SemEval-2017.

Further, Figure \ref{loss} demonstrates the descending curve of the training loss of different autoencoders. It shows that our attention autoencoders are more effective and robust than the RNN autoencoders in reconstructing the input sequences. In Table \ref{table case}, we present two examples from the development set to compare the reconstructed questions by RNN autoencoders and our attention autoencoders.

\begin{table}
\centering
\begin{tabular}{lp{0.72\columnwidth}}
\hline
Models & Questions \\
\hline
\hline
original & \emph{Where in Qatar is the best place for Snorkeling? I'm planning to go out next friday but don't know where to go.} \\
\hline
RNN & \emph{Where is the best place in Qatar for holiday? I'm going to go to sealine but don't know.} \\
\hline
attention & \emph{I'm planning to go out next friday but don't know where is the best place for Snorkeling in Qatar?} \\
\hline
\hline
original & \emph{Can anyone tell me where I can buy some indoor plants; besides the supermarket. Are there any nurseries in Doha? Also; are there any photography competitions in Doha? Thanks in advance} \\
\hline
RNN & \emph{Can anyone tell me where I can find some good outdoor plants; are there? ? ? are the only available in Doha; any other place in Doha? Thank you} \\
\hline
attention & \emph{Can anyone tell me where I can buy some indoor plants in Doha? Are there any nurseries; besides the indoor plants in Doha? Also; are there? Thanks in advance} \\
\hline
\end{tabular}
\caption{\label{table case} Comparison of reconstructed questions in RNN autoencoders and attention autoencoders.}
\end{table}

\section{Related Work}
\label{sect:related}

With the flourishing of CQA forums, research on question retrieval has attracted much attention. The studies that are close to our work can be roughly classified into the following categories.

\noindent{\bf Language model for information retrieval}. This kind of methods \cite{zhai2004study} computes the similarity based on the weights of the matching text terms between questions. \citeauthor{zhang2016learning} \shortcite{zhang2016learning} interpret language model methods from an embedding perspective. A key challenge of this kind of models is the lexical gap between new queries and existed questions.

\noindent{\bf Translation models}. In these models, the similarity function between questions is defined as the translation probability from a question to another one. \citeauthor{xue2008retrieval} \shortcite{xue2008retrieval} train a translation model from question-answer pairs. \citeauthor{zhou2011phrase} \shortcite{zhou2011phrase} propose a phrase-based translation model that can capture contextual information by treating phrases as a whole. A key limitation of these works is that they assume questions and answers are parallel texts and are semantic equivalent, which is hardly true in reality.

\noindent{\bf Topic Models}. In these methods, the similarity between questions is defined in the latent topic space \cite{ji2012question,zhang2014question}.

\noindent{\bf Deep learning based strategies}. Recently, various neural network architectures are applied to model question-question similarities. \citeauthor{das2016mirror} \shortcite{das2016mirror} propose a deep structured topic model, which first retrieve similar questions in the latent topic vector space and then reranking them using a deep layered semantic model. They train the neural network on question-answer pairs. \citeauthor{das2016together} \shortcite{das2016together} propose the Siamese Convolutional Neural Network for CQA (SCQA), which consists of twin convolutional neural networks with shared parameters. Our model is also under the flag of deep learning methods, but the difference is that our model is totally unsupervised and is more flexible.

\noindent{\bf SemEval campaign}. SemEval organizes a series of evaluations on CQA, SemEval-2015 Task 3 \cite{nakov2015semeval}, SemEval-2016 Task 3 \cite{nakov2016semeval} and SemEval-2017 Task 3 \cite{nakov2017semeval}, which provide benchmark datasets for evaluating different strategies on CQA. These tasks provide training data and so are in supervised setting \cite{franco2016uh,romeo2016neural,barron2016selecting}. We conduct experiments on this benchmark dataset, and compare our unsupervised model with state-of-the-art supervised methods.

\noindent{\bf Autoencoders}. Another relevant topic is autoencoders. Autoencoders is an unsupervised network method to learn latent representations that retain useful features to effectively reconstruct the inputs, which has been successfully applied in sentiment analysis, textual representations and so on. \citeauthor{li2015hierarchical} \shortcite{li2015hierarchical} employ hierarchical autoencoders to reconstruct paragraph representations. \citeauthor{makhzani2015adversarial} \shortcite{makhzani2015adversarial} propose the adversarial autoencoders, which apply generative adversarial networks (GAN) to perform variational inference.

\section{Conclusion}
\label{sect:conclusion}

In this paper, we present an unsupervised framework RAMN for question retrieval in CQA. The attention autoencoders, which are pre-trained using large scale unlabeled data in CQA archives, are applied to generate question representations. Further, we apply lexical mismatch information to effectively capture the surface matching between two questions. The final matching score is computed based on question representations, lexical mismatching information and the initial rank produced by a search engine. Our model enjoys the powerful matching capability of the deep semantic representations and at the same time captures the surface lexical matching. We conducted experiments at SemEval-2016 and SemEval-2017 CQA datasets. Our unsupervised model outperforms the winner system and is comparable with state-of-the-art methods which are all supervised. In future work, we will try to use the attention autoencoders for learning generic paragraph representations.

\section{Acknowledgments}
\label{sect:acknowledgments}
We thank Wei Li and Yixiu Wang for valuable comments and suggestion. This work is supported by National High Technology Research and Development Program of China (2015AA015403) and National Natural Science Foundation of China (61773026, 61371129). The corresponding author of this paper is Yunfang Wu.

\bibliography{Zhang}
\bibliographystyle{aaai}
\end{document}